\documentclass[keynote]{sigirforum}
\def\pubissue{Vol. 58 No. 1 -- June 2024}

\def\keynotedate{26 March 2024}

\begin{document}
% USE TITLE CASE FOR THE TITLE
\title{Shaping the Future of  Endangered and Low-Resource Languages  Our Role in the Age of LLMs: A Keynote at ECIR 2024}

\authors{
\author[Josiane.Mothe@irit.fr]{Josiane Mothe}{IRIT, UMR5505 CNRS, UT2J, Univ. de Toulouse}{France}
}

\maketitle 
\begin{abstract}
Isidore of Seville is credited with the adage that \textit{it is language that gives birth to a people, and not the other way around}, underlining the profound role played by language in the formation of cultural and social identity. Today, of the more than 7100 languages listed, a significant number are endangered. Since the 1970s, linguists, information seekers and enthusiasts have helped develop digital resources and automatic tools to support a wide range of languages, including endangered ones. The advent of Large Language Model (LLM) technologies holds both promise and peril. They offer unprecedented possibilities for the translation and generation of content and resources, key elements in the preservation and revitalisation of languages. They also present threat of homogenisation, cultural oversimplification and the further marginalisation of already vulnerable languages. The talk this paper is based on has proposed an initiatory journey, exploring the potential paths and partnerships between technology and tradition, with a particular focus on the Occitan language. Occitan is a language from Southern France, parts of Spain and Italy that played a major cultural and economic role, particularly in the Middle Ages. It is now endangered according to UNESCO. The talk critically has examined how human expertise and artificial intelligence can work together to offer hope for preserving the linguistic diversity that forms the foundation of our global and especially our European heritage while addressing some of the ethical and practical challenges that accompany the use of these powerful technologies. This paper is based on the keynote I gave at The 46th European Conference on Information Retrieval (ECIR 2024). As an alternative of reading this paper, you can find the video at \url{https://youtu.be/rwdlD7-7daY}.
\end{abstract}

\section{Why keep a language alive?}

Maybe one day will we be equipped with some device so that one can speak their mother tongue and others hear a perfect translation in their mother tongue. As a French, I can speak politics, secularism, food, my culture and every one understands what I mean. Interpreters disappeared but the diverse languages are alive as well as the culture they convey. But what is nowadays current situation.

According to François Gros-Jean, psycholinguistics, half of the population is bilingual at different levels~\citep{grosjean2013bilingualism}. And we use different languages in different situations. My mother tongue is French, it is my every day and all occasions language, I have been educated in French. Gascon, one of the Occitan language variants, is the one I like to speak with some friends, it was my parents’ mother tongue. 
English is the language I use mainly for research. At work, we speak a variant of English, the Globish with non-French native speakers. 
Indeed English serves as an international link language everywhere. 

%And, although there are now tools that  allow one to record a video in one language and automatically translate it into other major languages, it does not work yet for low resource languages like Gascon. 

With regard to English, there are almost 400 million first language speakers and 1.5 to 2 billion speakers around the world\footnote{\url{https://en.wikipedia.org/wiki/List_of_languages_by_total_number_of_speakers} Visited on March 2024}. 
So, it is natural that English was Sprint N°1 in our research in Information Retrieval (IR) and Natural Language Processinf (NLP). But have not we put other languages to the side? And can we make up for them now with LLMs?

According to Ethnologue\footnote{\url{https://www.ethnologue.com} Visited on March 2024}, there are more than 7100 languages, but 40\% of them are endangered. A language becomes endangered when parents do not speak their languages to their children but rather a more dominant language. In France, parents  used to speak French to their children rather than their mother tongue, as my parents did for me, because they wanted their children to be educated, to succeed in their life and thought that French was a better option than Gascon, as they were told.

But languages are not just a way to communicate. To see for yourself,  I encourage the reader to watch the video recorded during  the 2023 edition of the "festival des Bandas" in Condon~\footnote{\url{https://www.youtube.com/@festivalbandas5788}. This festival occurs every year the 2nd week of May.}. 

Indeed, the roles of languages are many, and not just to communicate.
It is part of our identity. It is used for education, influence and power, cultural transmission.

''It is language that gives birth to a people, and not the other way around.'' This adage is credited to San Isidoro de Sevilla and underlines the profound role languages play in our cultural and social identity. 
As computer scientists, we should not only consider the text we work on, but also what texts convey.

Languages are  our cultural heritage: a language is the repository of a community. Languages can help in understanding History. Like our historical monuments for example, where hardly anyone questions the need to preserve them. 
A language reflects the way its people understand the world. We all know that Inuits have many words for snow and ice. And that French is the most romantic language in the world. We do not want this diversity in perceiving the world to disappear, don’t we?
For some of us, it is thus only natural to want to preserve language diversity. For others, not so much. And, in History, imposing a language has served to unify and dominate diverse peoples.

Many seem to be aware that something important is at stake with LLMs.
March 2023, Iceland was very proud that Icelandic language was chosen for the developemnt on GPT4\footnote{\url{https://www.government.is/diplomatic-missions/embassy-article/2023/03/14/Head-start-for-Icelandic/}}.
An LLM for Croatian, Bosnian, Serbian, Montenegrin languages was announced late 2023~\footnote{\url{https://www.linkedin.com/posts/aleksagordic_first-ever-7-billion-parameter-hbs-llm-croatian-activity-7133414124553711616-Ep5J}}.
Taiwan is developing their LLMs\footnote{\url{https://english.cw.com.tw/article/article.action?id=3614}}.
Nora, an initiative with about 20 universities and research institutes, could be soon the LLM for Norwegian\footnote{\url{https://www.nora.ai/news/2023/the-nora.llm-project---large-language-models-as-in.html}}. 

One of the reasons to develop language-centric LLMs is to avoid artificial intelligence divide, and because lower abilities on smaller languages have been observed.
But it is also to preserve the distinctive character of the different languages; the texts LLM-based applications produce should be consistent with the culture and values of the people who speaks that language. 
The problem is that LLMs need a lot of data and this is a unique problem of low resource languages, they have not! In reality, low resource languages cover very different situations.

\section{Working on low-resource languages}
\citet{liu2022not} depict three different situations for low-resource languages\footnote{The authors considered hypothetical languages, we have adopted their ideas and terminology here, but consider them to be categories of languages rather than languages.}. 
All are low-resource languages, here we  which means there are less digital resources or NLP automatic tools in comparison to high resource languages. 
ELEPHANT languages, like Amharic, have 10 or more million of L1 speakers and millions of L2 speakers. An Elephant language is  institutional, used in all occasions; it is standardised. It is vital and is supported by multiple tools for NLP. The applications  users may need can be spoken language applications, information retrieval systems, mobile applications. To develop these applications, it is relatively easy to access to digital material and crowd sourcing platforms. There is an interest from industry and thus money can be found.
COYOTE languages, like Occitan is a completely different story. A COYOTE language is an endangered language. Applications should rather be to support the use of the language, language preservation, documentation and instruction. The IR community can help here, although it will not be easy to find language experts, there are maybe a few L1, elders and some L2 speakers. No interest from the industry, no money. 
The authors also mention OCELOT languages which situation is in between. 

A low resource or less resource language does not mean an endangered language but endangered languages are generally low-resource ones.

Occitan is now one of the endangered languages. Occitan is a Romance language. It was vital until the 14th century, Troubadours –artists- being well known for their “fin amor” courtly love style. Then, Occitan declined for many reasons and the revival efforts came late, just a few decades ago bilingual schools opened\footnote{\url{https://www.univ-montp3.fr/uoh/occitan/une_histoire/co/module_occitan_histoire_32.html}}.

%\section{Working on low-resource and endangered languages}

\begin{figure}
  \centering
   \includegraphics[scale=0.8]{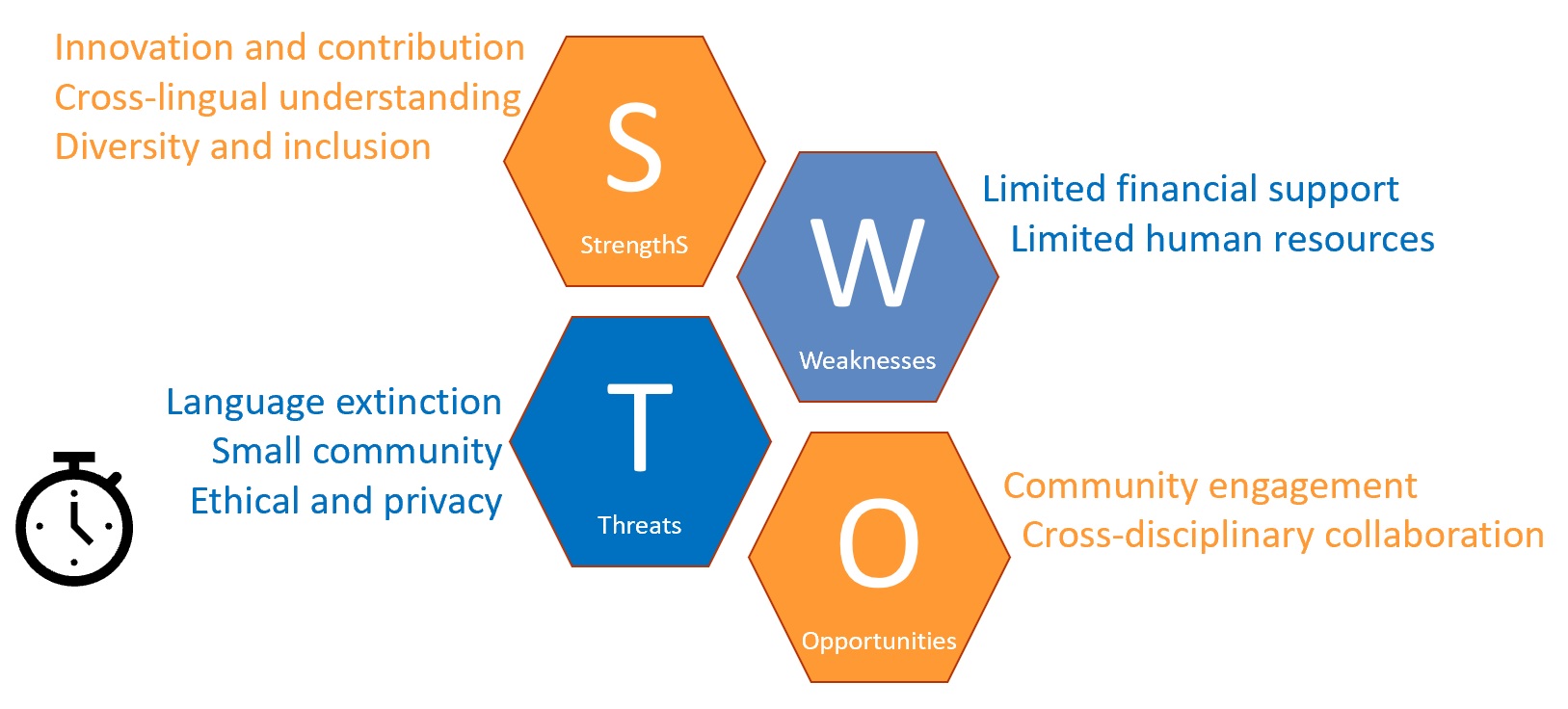}
    \caption{SWOT diagram on working on low-resource languages. } 
\label{Fig:SWOT}
\end{figure}
Figure~\ref{Fig:SWOT} is a SWOT diagram regarding the study of low-resource and endangered languages from a researcher’s point of view.

With regard to strengths, we can definitely contribute and innovate, more than on English, for which we have already done a lot. Also, understanding less studied languages may help us understand better higher resource languages and their variants. It also favours diversity and inclusion, with new researchers being involved like researchers for whom these low resource languages matter. 

With regard to weaknesses, there are limited financial support and limited human resources. We can probably positively influence on this.

With regard to threats, we have to be quick because some languages are disappearing and the communities we can rely on are small. We also have to convince them to work with us. 

Finally, with regard to opportunities, you can find very engaged people, who can help in data annotation and generation.  For those who like to work in a multi-disciplinary environment, this is also an opportunity.

We can work on shared tasks and I want here to recall the importance of shared tasks, if it was needed. Thanks to share tasks, we learnt to work in a controlled environment, in a scientific way, with reproducible methods and well acknowledged evaluation metrics. This is now the norm for more than 30 years in IR. 

TREC~\footnote{Text Retrieval Conference (\url{trec.nist.gov})} started with English, then with multi-lingual or cross-lingual IR, English being central, as one of the languages of the pairs.
But  it is just a few years ago that TREC really started to consider  low-resource languages. NeuCLIR track 2022 in TREC includes Persian as a medium-resource language~\citep{lawrie2024overview}. Others are FIRE with Urdu language~\citep{10.1145/3503162.3505240} or CIRAL for African languages~\citep{10.1145/3632754.3633076}. 
In NLP, it is worth  mentioning the conference on machine translation and its low-resource language track~\citep{pal-etal-2023-findings}. 

\section{LLMs - New opportunities for low-resource and endangered languages}
LLMs opened new opportunities for low-resource languages, let us have a look on what makes the LLMs quite unique: prompting, fine-tuning and Retrieval Augmented Generation.

One important step toward multi-lingual LLMs is cross-lingual word embeddings. In mono-lingual LLMs, the power of word embbedings lies in that two words with similar meanings are represented close to each other in the semantic space~\citep{mikolov2013exploiting,ruder2019survey}. Likewise, in multi-lingual embeddings, words that have similar meanings in different languages will be close to each other. That was an important milestone for many multilingual applications \citep{ruder2019survey}.

\subsection{Prompt engineering}
To leverage the knowledge that has been encoded in an LLM, you can prompt it. 
And it works impressively well, even if the encoding was from scarce data. 

\begin{figure}
  \centering
   \includegraphics[scale=0.8]{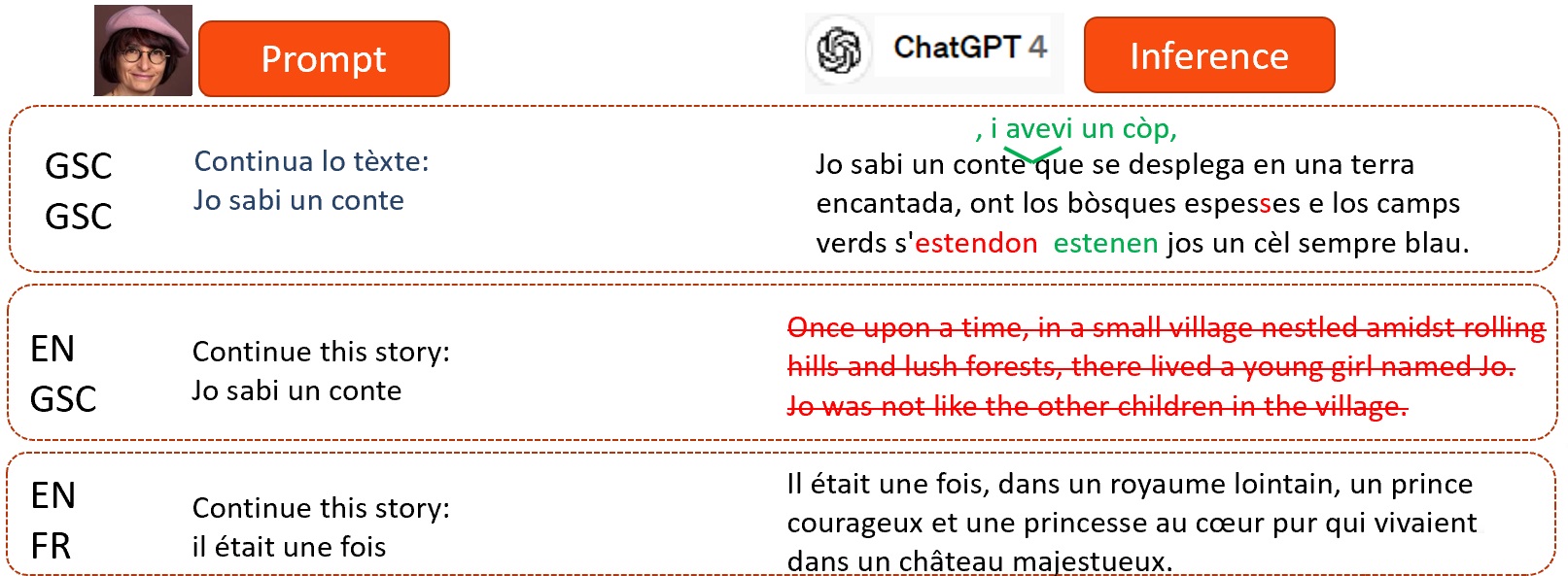}
    \caption{Auto-regressive prompting in different languages. Errors are highlighted in red and in green is a more appropriate answer.} 
\label{Fig:PromptRegressive}
\end{figure}
Figure~\ref{Fig:PromptRegressive} shows the results on Chat-GPT4 when using auto-regressive prompting, asking for continuing a text in different languages. Here, we consider what the LLM has been trained for: predicting the next token. 
First, when prompted  in Gascon (first row in Figure\ref{Fig:PromptRegressive}), it answers almost appropriately. GPT4 has seen some of the language, although we do not know exactly where and what, nor if it can distinguish the variants of the occitan language.

It does not work when asked in English to continue a text in Gascon, mixing the 2 languages in the prompt is not straightforward (2nd row in the Figure). Here, English overweight Gascon. Which is not the case when using two high resource languages, for example when mixing English and French works (3rd rwo in the Figure). For low resource languages,  more effort in prompt engineering is needed \citep{zhang2024teaching}. That also means we have to be careful when simulating low-resource  environments;  simulations can be biased as it is difficult to  know exactly the data the LLM has seen, even for open ones (language identification can be wrong too).

Likewise when using retrieval based prompting, the results strongly depend on the language used in the prompt. For example, when asked for stories for children in Occitan (Occitan being more generic than Gascon), when asked in English, the system fails: it does not know any Occitan stories nor where to find them. Querying in French, the answer is perfect, the system returns three appropriate links and descriptions. When  prompted in Occitan, it just provides some hits that could help to find such stories but does not provide any links.

So here, the lesson learnt is that we have to know on data used during training, which is not obvious, even for open LLMs. And yes, French libraries are more likely to sell books in Occitan. And no, prompting in English is not always the best.

We could also  mention few-shot prompting to instruct the LLM to perform tasks without extensive examples; this is very useful for low-resource languages for example on specific data annotation tasks such as language identification or sentiment analysis.
Also, chain of though could be designed for more complex tasks or to help the LLM to apply knowledge it learnt from high-resource languages to the target low-resource language. Sheila Teo shared how she won the first Singapore competition on prompt engineering~\citep{Sheila2023}. \citet{liu2023pre} wrote a survey on prompting methods for NLP.

\subsection{Fine tuning}
Fine tuning  is also very important for LLMs. \citet{artetxe-schwenk-2019-massively} want LLMs to be language-agnostic. Their model saw 93 languages. By learning tasks from English annotated data, it can be applied on the target task to any of the 93 languages without additional fine-tuning.

On machine translation task, \citet{zoph-etal-2016-transfer} introduced a transfer learning method where a model trained on a high-resource language pair (the parent model, here French to English) transfers knowledge to a low-resource language pair (the child model for example Uzbek to English).
The model is optimised by fixing certain parent model parameters (for example keeping the embeddings for English) and allowing others to be fine-tuned by the child model (the embedding from French are first copied to become Uzbek embedding and then Uzbek embedding are tuned). 

Instruction fine-tuning where the model is trained on instructions plus examples is also promising for low-resource languages~\citep{wei2021finetuned}, the more that the model can be self-instructed by using some texts~\citep{wang2022self}.

\subsection{Retrieval augmented generation and knowledge integration}

When queries extend beyond the model’s training data; which is likely to be the case for low-resource languages, LLMs may have hallucination. Retrieval Augmented Generation, which integrates external retrieved data into the generative process, can mitigate this~\citep{lewis2020retrieval}. The advantage is that it can be used at inference time. 
But in low-resource environment, the question may be which collection to search from? 

To provide additional information, knowledge graphs (KG) may be a better solution. KGs provide structured knowledge that LLMs lack of. Thus, combining them in an integrated framework is promising~\citep{pan2024unifying}.
For low-resource languages, KG or external knowledge can be dictionaries or grammar rules, or any other kinds of resource. 
But LLMs can also help in building KGs and for low-resource languages, LLMs can help building language representations and description for languages that lack of.~\citep{tan2024large}

\section{Users' perspective}
In our work with Addis-Abeba University in Ethiopia, we wanted to contribute changing the low-resource situation of Amharic,  specially for IR.  Most of the existing data and tools were not open data. We have developed open data in Amharic: an IR adhoc collection~\citep{yeshambel20202airtc}, morphologically annotated corpora~\citep{yeshambel2021morphologically}, embeddings~\citep{info14030195}, ...
Most of the resources we developed can be found at \url{https://www.irit.fr/AmharicResources/}

On Occitan, data has been collected and a treebank developed by \citet{bras2016bateloc}. Treebanks are structured databases of natural language sentences annotated to show grammatical structures and relationships between words or phrases. Treebanks play a crucial role in linguistic analysis.
The Universal Dependencies (UD) project~\footnote{\url{https://universaldependencies.org/}} focuses on defining and using a uniform  representation for all the languages. This can facilitate comparison and research across different languages. UD is an open community effort with over 500 contributors in over 100 languages. 

Indeed, for endangered languages, it is crucial to document them. Manual annotation is very slow and automatic tools can be very helpful \citep{bernhard2020avenir}.
Even before LLMs, researchers used cross lingual annotation for annotating texts. That means they use tools developed on a close language and apply them to the target new language for a first automatic annotation. Then, they manually correct the annotations and use reinforcement learning to continuously improve the tools on the target language.

Something which is important for low resource languages is the media presence with up to date content. LLMs can help in generating content for different audiences on social media for exemple.

Machine translation is also a very important aspect.

\section{Machine translation}

\subsection{Data sets}

More and more  data sets are  open data. Some are just harvested data, with no or little curation and verification such as the common crawl~\footnote{\url{https://commoncrawl.org/}} or OPUS data~\citep{zhang2020improving}. Others, like the Universal Declaration of Human Rights (UDHR)~\footnote{\url{https://huggingface.co/datasets/udhr}} or religious books exist in many languages~\footnote{\url{https://data.world/datasets/bible}}, almost aligned. 

There are also important collective work corpora. 
The most popular maybe is FLORES-101~\citep{goyal2022flores}, extended to FLORES-200~\footnote{\url{https://paperswithcode.com/dataset/flores-200}}. FLORES-200 is a collection of English sentences translated into about 200 languages, and humanly-verified. 
AYA, which has been recently released, is a crowd-sourced collection of prompts and their completion~\citep{singh2024aya}, with more than 100 languages and more than 5 hundred millions of instances (although some are automatically translated). 

Some of these resources are used to train models, others to evaluate machine translation, so there are of very different nature, because for different purposes.

What has been found is LLM-based machine translation systems perform better when translating into English compared to translating from English (See Equation~\ref{Equ}).
\begin{equation}
    SS -> EN >> EN -> XX 
    \label{Equ}
\end{equation}

Here, $XX$ means one or several languages other than English (EN). That could mean English generation is easier. 
How much is it linked to the linguistic imperialism and abundance of data? It is just because we have focused more our energy on research in English? 

English has a relatively straightforward grammatical structure compared to many other languages, with less inflection and fewer grammatical gender considerations. On the other hand, many data collections for evaluation are translations from English to another language, making back-translation an easier task.

Experiments have shown the knowledge on the target language seems to have an impact. Decoding could be key here, even more than coding. However, LLMs have been trained on a large portion of English and human feedback on English. Is that also the reason why language similarity with English seems to be important?

The interrelation between the huge effort put on English and these results is not clear, neither how this generalises to other languages.

Also about $XX$ in Equation~\ref{Equ}: languages are many and different. I think we should consider them more than just numbers and maybe that is not appropriate to average the results obtained when translating on different languages into $XX$  (across languages). Languages deserve more respect! 

BLEU (BiLingual Evaluation Understudy) is one of the measures to evaluate machine translation~\citep{papineni2002bleu}, which is usually reported in between 0 and 100. In some studies, tables report BLEU measurements that are averaged across different languages so that a 3.4 for Armenian for example and a 66 for Scot Gaelic can contribute the same to the average. 

In IR, we also average results across  queries with MAP (Mean Average Precision) for example, were precision is averaged across queries. For those who know my work on query performance prediction~\citep{mothe2005linguistic,chifu2018query} and adaptive query processing~\citep{BigotEtAl2015,mothe2023selective}, you also know that I want to consider queries as not just one in many; but see what we could do to answer each of them. 

Another concern with LLMs is their output. They produce texts, but is there not a risk here that languages might lose more than they gain?
Is text generation sufficient for understanding a language and ensuring its transmission to future generations?

While examples are helpful for illustration, understanding also requires concepts and rules for generalization. Should we entrust this responsibility to LLMs?

LLMs can generate text in nearly any language, but does this imply that humans will truly understand and use the language correctly? What insights can humans gain from the text generated by these models? Will it be necessary for humans to relearn languages based solely on these texts?  In a way, isn't that what we are doing right now, trying to understand exactly how LLMs work? Furthermore, is it prudent to disregard the wealth of knowledge accumulated through other methods?

\section{New directions}
Never before systems have been so greedy. It seems that the bigger the smarter, and there is a tendency to train models with more data, more parameters, more modes (including text, image, video, sounds, data), more languages (some have been trained on more than 100 languages), more context (e.g. Google Gemini has now 1 M tokens context), more tasks (conversational, QA, machine translation, summarization).  
So, we want to have a universal machine, that can answer any task.
Now for other languages that are not yet included, this is tempting to reproduce what have been done for high resource languages; LLMs seem to open so many opportunities…
But, should we have a model that would include the 7 000 languages? no matter how big and expensive it is and how costly it is, also in terms of foot print?

Alternatively, we might consider the concept of degrowth in the development of such systems. Operating in low-resource environments necessitates a different approach: fewer parameters require less data to tune, as suggested by the so-called "chinchilla law."~\citep{hoffmann2022training} Consequently, smaller models not only require less computational power but also can efficiently handle specific, narrow tasks that meet users’ needs.

Mistral~\footnote{\url{https://mistral.ai/fr/news/mixtral-of-experts/}} has emerged as one of the pioneers in promoting this shift towards simpler, more sustainable models, possibly setting a trend for lightweight and open-access technologies. Techniques such as Parameter Efficient Fine Tuning are particularly valuable in situations with scarce labelled data, allowing for strategic decisions on which parts of the model to train and which to leave unchanged~\citep{ding2023parameter}.

Moreover, methods like Knowledge Distillation~\citep{west2021symbolic}, LLM pruning~\citep{ma2023llm}, and quantisation~\citep{xiao2023smoothquant} have been developed to enhance efficiency and facilitate the deployment of these models on portable devices or in specialised fields. Although these techniques were not originally intended for endangered or low-resource languages, they hold potential applications for these areas as well. 

Despite their diversity and number, endangered languages face the same significant challenges, including: a shortage of data, a lack of specialists, and a heightened risk of relying on flawed automatic translation tools for generating digital data. 
The communities also share similar needs, suggesting that a collaborative effort could potentially trigger a widespread positive impact on all endangered languages. Despite their small size, the consistency in their challenges makes a united approach viable and a butterfly effect on many of the endangered languages possible. Each community is small but, they face the same challenges, needs and concerns. 

From an ethical perspective, it is crucial to assure these communities that their linguistic data is safer under professional care than if managed solely by themselves. This requires transparent communication about our intentions and methodologies. The language preservation should extend beyond simple academic metrics such as BLEU scores.

Furthermore, our motivation for studying languages should transcend practical or immediate needs driven by factors such as war, cybersecurity threats, immigration, or economic considerations. Instead, we should also focus on the intrinsic value of languages, celebrating diversity and culture—which are core European values. The diversity within the English language alone merits research attention. Another layer of complexity in language preservation lies in that not all languages have a written form, for example  Occitan whistled language~\citep{biu2018enseignement}).

\section*{Acknowledgments}
I am thankful to colleagues and friends including Myriam Bras, Claude Chrisment, Bruno Menon, Aleksandra Miletic, Philippe Muller, Didier Peyrusqué, Catherine Stringer, Tilahun Yeshambel  who provided  expertise and lively discussions that greatly assisted the content of this keynote.
\bibliography{Mothe}
\end{document}